\documentclass[10pt, a4paper]{article}
\usepackage{lrec}
\usepackage{graphicx}
\usepackage{tabularx}
\usepackage{soul}

\usepackage{epstopdf}
\usepackage[utf8]{inputenc}

\usepackage{hyperref}
\usepackage{xstring}
\usepackage[T1]{fontenc}
\usepackage[normalem]{ulem}
\usepackage{color}

\newcommand{\secref}[1]{\StrSubstitute{\getrefnumber{#1}}{.}{}}

\title{LibriVoxDeEn: A Corpus for German-to-English Speech Translation \\ and German Speech Recognition}

\name{Benjamin Beilharz$^{\ast}$, Xin Sun$^{\ast}$, Sariya Karimova$^{\ast}$, Stefan Riezler$^{\dagger,\ast}$}

\address{$^{\ast}$Computational Linguistics \& $^{\dagger}$IWR\\
Heidelberg University, Germany \\
\small \tt \{beilharz,karimova,riezler\}@cl.uni-heidelberg.de,\\ 
\tt xin.sun@stud.uni-heidelberg.de}

\abstract{
We present a corpus of sentence-aligned triples of German audio, German text, and English translation, based on German audio books. The speech translation data consist of 110 hours of audio material aligned to over 50k parallel sentences. An even larger dataset comprising 547 hours of German speech aligned to German text is available for speech recognition. The audio data is read speech and thus low in disfluencies. The quality of audio and sentence alignments has been checked by a manual evaluation, showing that speech alignment quality is in general very high. The sentence alignment quality is comparable to well-used parallel translation data and can be adjusted by cutoffs on the automatic alignment score. To our knowledge, this corpus is to date the largest resource for German speech recognition and for end-to-end German-to-English speech translation.
\\ \newline 
}

\begin{document}

\maketitleabstract

\section{Introduction}

Direct speech translation has recently been shown to be feasible using a single sequence-to-sequence neural model, trained on parallel data consisting of source audio, source text and target text. The crucial advantage of such end-to-end approaches is the avoidance of error propagation as in a pipeline approaches of speech recognition and text translation. While cascaded approaches have an advantage in that they can straightforwardly use large independent datasets for speech recognition and text translation, clever sharing of sub-networks via multi-task learning and two-stage modeling \cite{WeissETAL:17,AnastasopoulosChiang:18,SperberETAL:19} has closed the performance gap between end-to-end and pipeline approaches. However, end-to-end neural speech translation is very data hungry while available datasets must be considered large if they exceed 100 hours of audio. For example, the widely used Fisher and Callhome Spanish-English corpus \cite{PostETAL:13} comprises 162 hours of audio and $138,819$ parallel sentences. Larger corpora for end-to-end speech translation have only recently become available for speech translation from English sources. For example, 236 hours of audio and $131,395$ parallel sentences are available for English-French speech translation based on audio books \cite{KocabiyikogluETAL:18,BerardETAL:18}. For speech translation of English TED talks, 400-500 hours of audio aligned to around $250,000$ parallel sentences depending on the language pair have been provided for eight target languages by \newcite{DiGangiETAL:19}. Pure speech recognition data are available in amounts of $1,000$ hours of read English speech and their transcriptions in the {LibriSpeech} corpus provided by \newcite{PanayotovETAL:15}.

When it comes to German sources, the situation regarding corpora for end-to-end speech translation as well as for speech recognition is dire. To our knowledge, the largest freely available corpora for German-English speech translation comprise triples for 37 hours of German audio, German transcription, and English translation \cite{StuekerETAL:12}. Pure speech recognition data are available from 36 hours \cite{Radeck-ArnethETAL:15} to around 200 hours \cite{BaumannETAL:18}. 

We present a corpus of sentence-aligned triples of German audio, German text, and English translation, based on German audio books. The corpus consists of 110 hours of German audio material aligned to over 50k parallel sentences. An even larger dataset comprising 547 hours of German speech aligned to German text is available for speech recognition. Our approach mirrors that of \newcite{KocabiyikogluETAL:18} in that we start from freely available audio books. The fact that the audio data is read speech keeps the number of disfluencies low. Furthermore, we use state-of-the-art tools for audio-text and text-text alignment, and show in a manual evaluation that the speech alignment quality is in general very high, while the sentence alignment quality is comparable to widely used corpora such as that of \newcite{KocabiyikogluETAL:18}, and can be adjusted by cutoffs on the automatic alignment score. To our knowledge, the presented corpus is to date the largest resource for German speech recognition and for end-to-end German-to-English speech translation.

\section{Overview}

In the following, we will give an overview over our corpus creation methodology. More details will be given in the following sections.

\begin{enumerate}
    \item Creation of German speech recognition corpus (see Section \secref{source_corpus})
	\begin{itemize}
		\item Data download
		\begin{itemize}
		    \item Download German audio books from \emph{LibriVox} web platform
		    \footnote{\url{https://librivox.org}}
		    \item Collect corresponding text files by crawling public domain web pages\footnote{\url{https://gutenberg.spiegel.de,http://www.zeno.org, https://archive.org}}
	    \end{itemize}
		\item Audio preprocessing
		\begin{itemize}
		    \item Manual filtering of audio pre- and postfixes 
		\end{itemize}
		\item Text preprocessing
		\begin{itemize}
		    \item Noise removal, e.g. special symbols, advertisements, hyperlinks
		    \item Sentence segmentation using \emph{spaCy}\footnote{\url{https://spacy.io/}}
	    \end{itemize}
	    \item Speech-to-text alignments
	        \begin{itemize}
		        \item Manual chapter segmentation of audio files
		        \item Audio-to-text alignments using forced aligner \emph{aeneas}\footnote{\url{https://github.com/readbeyond/aeneas}}
		        \item Split audio according to obtained timestamps using \emph{SoX}\footnote{\url{http://sox.sourceforge.net/}}
	        \end{itemize}
	\end{itemize}
    \item Creation of German-English Speech Translation Corpus (see Sections \secref{target_corpus} { } and \secref{corpus_filtering})
    
    \begin{itemize}
        \item Download English translations for German texts
		\item Text preprocessing (same procedure as for German texts)
		\item Bilingual text-to-text alignments
		\begin{itemize}
		    \item Manual text-to-text alignments of chapters
    		\item Dictionary creation using parallel DE-EN \emph{WikiMatrix}\footnote{\url{https://ai.facebook.com/blog/wikimatrix/}} corpus \cite{SchwenkETAL:19}
		    \item German-English sentence alignments using \emph{hunalign} \cite{VargaETAL:05}
		    \item Data filtering based on \emph{hunalign} alignment scores
	    \end{itemize}
	\end{itemize}
\end{enumerate}

\section{German Speech Recognition Data}
\label{source_corpus}

\subsection{Data Collection}

We acquired pairs of German books and their corresponding audio files starting from \emph{LibriVox}, an open source platform for people to publish their audio recordings of them reading books which are available open source on the platform \emph{Project Gutenberg}. German data were gathered in a semi-automatic way: The URL links were collected manually by using queries containing metadata descriptions to find German books with LibriVox audio and possible German transcripts. These were later automatically scraped using \emph{BeautifulSoup4}\footnote{\url{https://www.crummy.com/software/BeautifulSoup/bs4/doc/}} and \emph{Scrapy}\footnote{\url{https://scrapy.org/}}, and saved for further processing and cleaning. 
Public domain web pages crawled include \url{https://gutenberg.spiegel.de}, \url{http://www.zeno.org}, and \url{https://archive.org}. 

\subsection{Data Preprocessing}

We processed the audio data in a semi-automatic manner which included manual splitting and alignment of audio files into chapters, while also saving timestamps for start and end of chapters. We removed boilerplate intros and outros and as well as noise at the beginning and end of the recordings.

Preprocessing the text included removal of several items, including special symbols like *, advertisements, hyperlinks in [], <>, empty lines, quotes, - preceding sentences, indentations, and noisy OCR output.

German sentence segmentation was done using \emph{spaCy} based on a medium sized German corpus\footnote{\url{https://spacy.io/models/de\#de\_core\_news\_md}} that contains the TIGER corpus\footnote{\url{https://www.ims.uni-stuttgart.de/forschung/\\ressourcen/korpora/tiger.html}} and the  WikiNER\footnote{\url{https://dx.doi.org/10.1016/j.artint.2012.03.006}} datasets. Furthermore we added rules to adjust the segmenting behavior for direct speech and for semicolon-separated sentences.


\subsection{Text-to-Speech Alignment}

To align sentences to onsets and endings of corresponding audio segments we made use of \emph{aeneas} -- a tool for an automatic synchronization of text and audio. In contrast to most forced aligners, \emph{aeneas} does not use automatic speech recognition to compare an obtained transcript with the original text. Instead, it works in the opposite direction by using dynamic time warping to align the mel-frequency cepstral coefficients extracted from the real audio to the audio representation synthesized from the text, thus aligning the text file to a time interval in the real audio.

Furthermore, we used the maps pointing to the beginning and the end of each text row in the audio file produced with \emph{SoX} 
to split the audio into sentence level chunks. The timestamps were also used to filter boilerplate information about the book, author, speaker at the beginning and end of the audio file. 

Statistics on the resulting corpus are given in Table \ref{tab:source}. The corpus consists of 86 audio books, mostly fiction, comprising $547$ hours of audio, aligned to over $400,000$ sentences and over $4$M words.

\begin{table*}[!t]
\begin{center}
\begin{tabular}{|c|c|c|c|c|c|c|}
      \hline
      \#books & \#chapters & \#sentences & \#hours & \#words & sampling rate & resolution\\ 
      \hline
      86 & 1,556 & 419,449 & \textbf{547} & 4,082,479 & 22 kHz & 16 bit\\
      \hline
\end{tabular}
\caption{Statistics of German speech recognition corpus}
\label{tab:source}
 \end{center}
\end{table*}

\begin{table*}[!t]
\begin{center}
\begin{tabular}{|c|c|c|c|c|}
      \hline
      \#books & \#chapters & \#sentences & \#hours& \#words\\ 
      \hline
      19 & 365 & [DE] 53,168 & \textbf{133} & [DE] 898,676 \\
      & & [EN] 50,883 & & [EN] 989,768 \\
      \hline
\end{tabular}
\caption{German(DE)-English(EN) text-to-text alignment data}
\label{tab:pre}
 \end{center}
\end{table*}

\begin{table*}[!t]
\begin{center}
\begin{tabular}{|c|c|c|c|c|}
      \hline
      \#books & \#chapters & \#sentences & \#hours & \#words\\ 
      \hline
      19 & 365 & [DE] 50,427 & \textbf{110} & [DE] 860,369 \\  
      & & [EN] 50,883 & & [EN] 948,565 \\  
      \hline
\end{tabular}
\caption{German(DE)-English(EN) text-to-text alignment data after filtering}
\label{tab:post}
 \end{center}
\end{table*}

\section{German-to-English Parallel Text Data}
\label{target_corpus}

\subsection{Data Collection and Preprocessing}
In collecting and preprocessing the English texts we followed the same procedure as for the German source language corpus, i.e., we manually created queries containing metadata descriptions of English books (e.g. author names) corresponding to German books which then were scraped. The \emph{spaCy} model for sentence segmentation used a large English web corpus\footnote{\url{https://spacy.io/models/en\#en\_core\_web\_lg}}. See Section \secref{source_corpus} { } for more information.

\subsection{Text-to-Text Alignment}

To produce text-to-text alignments we used \emph{hunalign} with a custom dictionary of parallel sentences, generated from the \emph{WikiMatrix} corpus. Using this additional dictionary improved our alignment scores. Furthermore we availed ourselves of a realign option enabling to save a dictionary generated in a first pass and profiting from it in a second pass. The final dictionary we used for the alignments consisted of a combination of entries of our corpora as well as the parallel corpus \emph{WikiMatrix}. For further completeness we reversed the arguments in \emph{hunalign} to not only obtain German to English alignments, but also English to German. These tables were merged to build the union by dropping duplicate entries and keeping those with a higher confidence score, while also appending alignments that may only have been produced when aligning in a specific direction. 

Statistics on the resulting text alignments are given in Table \ref{tab:pre}.

\section{Data Filtering and Corpus Structure}
\label{corpus_filtering}

\subsection{Corpus Filtering}

A last step in our corpus creation procedure consisted of filtering out empty and  incomplete alignments, i.e., alignments that did not consist of a DE-EN sentence pair. This was achieved by dropping all entries with a \emph{hunalign} score of -0.3  or below. 
Table \ref{tab:post} shows the resulting corpus after this filtering step.

Moreover, many-to-many alignments by \emph{hunalign} were re-segmented to source-audio sentence level for German, while keeping the merged English sentence to provide a complete audio lookup.
The corresponding English sentences were duplicated and tagged with \texttt{<MERGE>} to mark that the German sentence was involved into a many-to-many alignment. 

The size of our final cleaned and filtered corpus is thus comparable to the cleaned Augmented LibriSpeech corpus that has been used in speech translation experiments by \newcite{BerardETAL:18}.

Statistics on the resulting filtered text alignments are given in Table \ref{tab:post}.

\subsection{Corpus Structure}

Our corpus is structured in following folders:
\begin{itemize}
    \item[de]
    \begin{itemize}
        \item contains German text files for each book
    \end{itemize}
    \item[en]
    \begin{itemize}
        \item contains English text files for each book
    \end{itemize}
    \item[audio]
    \begin{itemize}
        \item alignment maps produced by \emph{aeneas}
        \item sentence level audio files
    \end{itemize}
    \item[tables]
    \begin{itemize}
        \item text2speech, a lookup table for speech alignments
        \item text2text, a lookup table for text-to-text alignments
    \end{itemize}
\end{itemize}

Further information about the corpus and a download link can be found here: \url{https://www.cl.uni-heidelberg.de/statnlpgroup/librivoxdeen/}.


\begin{table*}[!t]
\begin{center}
\begin{tabular}{|c|c|c|c|}
      \hline
      \textbf{Bin} & \textbf{\emph{hunalign} confidence (avg) } & \textbf{audio-text alignment (max 3)} & \textbf{text-text alignment (max 5)}\\
      \hline
      Low & 0.17 & 2.73 & 3.43  \\
      \hline
      Moderate & 0.59 & 2.65 & 3.63 \\
      \hline
      High & 1.06 & 2.71 & 4.35 \\
      \hline
      \textbf{Average} & \textbf{0.61} & \textbf{2.69} & \textbf{3.80}\\
      \hline
\end{tabular}
\caption{Manual evaluation for audio-text and text-text alignments, averaged over 90 items and two raters}
\label{tab:res}
 \end{center}
\end{table*}

\section{Corpus Evaluation}

\subsection{Human Evaluation}

For a manual evaluation of our dataset, we split the corpus into three bins according to ranges $(-0.3,0.3]$, $(0.3,0.8]$ and $(0.8,\infty)$ of the \emph{hunalign} confidence score (see Table \ref{tab:bins}). 

\begin{table}[!h]
\begin{center}
\begin{tabular}{|c|c|c|c|}
      \hline
      & \textbf{Low} & \textbf{Moderate} & \textbf{High} \\
      \hline
      \textbf{Bin} & $-0.3 < x \leq 0.3$ & $0.3 < x \leq 0.8$ & $0.8 < x$\\
      \hline
\end{tabular}
\caption{Bins of text alignment quality according to \emph{hunalign} confidence score}
\label{tab:bins}
 \end{center}
\end{table}

The evaluation of the text alignment quality was conducted according to the 5-point scale used in \newcite{KocabiyikogluETAL:18}:
\begin{enumerate}
    \item[1]Wrong alignment
    \item[2]Partial alignment with slightly compositional translational equivalence
    \item[3]Partial alignment with compositional translation and additional or missing information
    \item[4]Correct alignment with compositional translation and few additional or missing information
    \item[5]Correct alignment and fully compositional translation
\end{enumerate}

The evaluation of the audio-text alignment quality was conducted according to the following 3-point scale: 
\begin{enumerate}
    \item[1] Wrong alignment
    \item[2] Partial alignment, some words or sentences may be missing
    \item[3] Correct alignment, allowing non-spoken syllables at start or end
\end{enumerate}

The evaluation experiment was performed by two annotators who each rated 30 items from each bin, where 10 items were the same for both annotators in order to calculate inter-annotator reliability. 

\subsection{Evaluation Results}

Table \ref{tab:res} shows the results of our manual evaluation. 
The audio-text alignment was rated as in general as high quality. The text-text alignment rating increases corresponding to increasing \emph{hunalign} confidence score which shows that the latter can be safely used to find a threshold for corpus filtering. Overall, the audio-text and text-text alignment scores are very similar to those reported by \newcite{KocabiyikogluETAL:18}.

The inter-annotator agreement between two raters was measured by Krippendorff's $\alpha$-reliability score \cite{Krippendorff:13} for ordinal ratings. The inter-annotator reliability for text-to-text alignment quality ratings scored 0.77, while for audio-text alignment quality ratings it scored 1.00.

\subsection{Examples}

In the following, we present selected examples for text-text alignments for each bin. A closer inspection reveals properties and shortcomings of \emph{hunalign} scores which are based on a combination of dictionary-based alignments and sentence-length information. 

Shorter sentence pairs are in general aligned correctly, irrespective of the score (compare examples with scores $0.30$, $0.78$, $1.57$, and $2.44$ below). Longer sentences can include exact matches of longer substrings, however, they are scored based on a bag-of-words overlap (see the examples with scores $0.41$ and $0.84$ below). This heuristics works well for examples at the low and high end of the range of \emph{hunalign} scores (scores $-0.06$ and $0.02$ indicate bad alignments, scores higher than $0.75$ correspond to relatively good alignments).

\begin{itemize}
    \item[-0.06]
    \begin{itemize}
        \item[DE]Schigolch Yes, yes; und mir träumte von einem Stück Christmas Pudding.	
        \item[EN]She only does that to revive old memories. ~~~ LULU.
    \end{itemize}
    \item[0.02]
    \begin{itemize}
        \item[DE]Und hätten dreißigtausend Helfer sich ersehn.
        \item[EN]And feardefying Folker shall our companion be; He shall bear our banner; better none than he.
    \end{itemize}
    \item[0.30]
    \begin{itemize}
        \item[DE]Kakambo verlor nie den Kopf.
        \item[EN]Cacambo never lost his head.
    \end{itemize}
    \item[0.41]
    \begin{itemize}
        \item[DE]Es befindet sich gar keine junge Dame an Bord, versetzte der Proviantmeister.
        \item[EN]He is a tall gentleman, quiet, and not very talkative, and has with him a young lady — There is no young lady on board, interrupted the AROUND THE WORLD IN EIGPITY DAYS. purser..
    \end{itemize}
    \item[0.75]
    \begin{itemize}
	\item[DE]Ottilie, getragen durch das Gefühl ihrer Unschuld, auf dem Wege zu dem erwünschtesten Glück, lebt nur für Eduard.
	\item[EN]Ottilie, led by the sense of her own innocence along the road to the happiness for which she longed, only lived for Edward.
    \end{itemize}
    \item[0.78]
    \begin{itemize}
        \item[DE]Was ist geschehen? fragte er.
        \item[EN]What has happened ? he asked.
    \end{itemize}
    \item[0.84]
    \begin{itemize}
        \item[DE]Es sind nun drei Monate verflossen, daß wir Charleston auf dem Chancellor verlassen, und zwanzig Tage, die wir schon auf dem Flosse, von der Gnade der Winde und Strömungen abhängig, verbracht haben!
        \item[EN]JANUARY st to th.More than three months had elapsed since we left Charleston in the Chancellor, and for no less than twenty days had we now been borne along on our raft at the mercy of the wind and waves.
    \end{itemize}
    \item[1.57]
    \begin{itemize}
        \item[DE]Charlotte stieg weiter, und Ottilie trug das Kind.
        \item[EN]Charlotte went on up the cliff, and Ottilie carried the child.
    \end{itemize}
    \item[2.44]
    \begin{itemize}
        \item[DE]Fin de siecle, murmelte Lord Henry.
        \item[EN]Fin de siecle, murmured Lord Henry.
    \end{itemize}
    
\end{itemize}

\section{Conclusion}

\begin{table*}[!t]
\begin{center}
\begin{tabular}{|c|c|c|p{3.7cm}|p{3.7cm}|c|c|}
     \hline
     book & audio & score & de\_sentence & en\_sentence & \#w\_de & \#w\_en\\ 
     \hline
     18.undine & 00001-undine10.wav & 0.63 & Ja, als er die Augen nach dem Walde aufhob, kam es ihm ganz eigentlich vor, als sehe er durch das Laubgegitter den nickenden Mann hervorkommen. & Indeed, when he raised his eyes toward the wood it seemed to him as if he actually saw the nodding man approaching through the dense foliage. & 25 & 26\\
     \hline
\end{tabular}
\end{center}
\caption{Example entry in LibriVoxDeEn, listing name of book file, name of audio file, hunalign score, German sentence, aligned English sentence, number of words in German sentence, number of words in English sentence.}
\label{tab:entry}
\end{table*}

We presented a corpus of aligned triples of German audio, German text, and English translations for speech translation from German to English. An example entry is given in Table \ref{tab:entry}. The audio data in our corpus are read speech, based on German audio books, ensuring a low amount of speech disfluencies. The audio-text alignment and text-to-text sentence alignment was checked to be of high quality in a manual evaluation.  A cutoff on a sentence alignment quality score allows to filter the text alignments further, resulting in a clean corpus of $50,427$ German-English sentence pairs aligned to 110 hours of German speech. A larger version of the corpus, comprising 547 hours of German speech and high-quality alignments to German transcriptions is available for speech recognition. 

\section{Acknowledgments}

We would like to thank the anonymous reviewers for their feedback. 
The research reported in this paper was supported in part by the German research foundation (DFG) under grant RI-2221/4-1.

\section{Bibliographical References}
\bibliographystyle{lrec}
\bibliography{references}

\end{document}